\pdfoutput=1

\documentclass[11pt]{article}

\usepackage[final]{acl}

\usepackage{times}
\usepackage{latexsym}

\usepackage[T1]{fontenc}

\usepackage[utf8]{inputenc}

\usepackage{microtype}

\usepackage{inconsolata}

\usepackage{amsmath}
\usepackage{amssymb}
\usepackage{booktabs}
\usepackage{graphicx}
\graphicspath{{figures/}}
\usepackage{multirow}
\usepackage{xspace}
\usepackage{float}

\usepackage{tabularx}
\usepackage[breakable, most]{tcolorbox}
\usepackage{xcolor}
\usepackage{longtable}
\usepackage{qrcode}
\usepackage{placeins}
\definecolor{LightBlue}{HTML}{3389f8}
\usepackage{amsthm}
\newtheorem{proposition}{Proposition}

\newcommand{\mabpd}{\textsc{MABPD}\xspace}

\title{MABPD: Multi-Agent Bias Probing \& Detection\\via Structured Argument Debate}

\author{
  Garvit Joshi, Stavya Dhyani, Jasmine, Arun Chauhan \\
  Graphic Era University, Dehradun, India
}

\begin{document}
\maketitle

\begin{abstract}
Media bias in news articles operates through subtle linguistic cues---loaded language, selective framing, and strategic omission---that resist single-model detection and have traditionally required large annotated corpora for supervised training.
We ask whether \emph{structured multi-agent deliberation} can serve as a principled, training-free alternative to supervised classification for this task.
We introduce \mabpd (\textbf{M}ulti-\textbf{A}gent \textbf{B}ias \textbf{P}robing \& \textbf{D}etection), a pipeline in which three specialized LLM agents analyze an article from complementary perspectives and resolve disagreements through a \emph{Structured Argument Debate} (SAD) protocol.
SAD implements a domain-motivated asymmetric burden of proof---biased claims without grounded textual evidence carry zero weight---combined with role-weighted voting and post-consensus verification, replacing task-specific supervised decision boundaries with explicit deliberative structure.
Ablation confirms that this structured deliberation, not mere agent parallelism, drives performance: removing the debate module reduces F1 by up to 10.6 points.
On the BABE benchmark (4{,}121 expert-annotated sentences), \mabpd achieves \textbf{83.4\% macro F1} on the held-out test split---within 0.7 percentage points (pp) of the supervised SOTA (MAGPIE, 84.1\% macro F1; \citealp{Horych2024magpie})---\textbf{without any task-specific training or threshold tuning on annotated data}.
Cross-dataset evaluation on the SemEval 2019 HyperPartisan corpus (644 articles) yields 75.0\% zero-shot accuracy, within 7.2~pp of the supervised SOTA accuracy (82.2\%; \citealt{Kiesel2019hyperpartisan}), confirming transfer across annotation regimes.
We release the full pipeline and evaluation code.\footnote{Code: \url{https://github.com/Subaru-5999/MABPD}}
\end{abstract}

\section{Introduction}
\label{sec:intro}

Media bias in news articles influences public opinion on a massive scale, yet reliable computational detection remains elusive.
Unlike factual misinformation, media bias operates through \emph{how} information is presented---via loaded language, selective framing, ideological labeling, and strategic omission of counter-evidence \citep{Wessel2023mbib,Fan2019babe}.
A single sentence may be biased through loaded word choice while appearing factually accurate, making binary classification insufficient without deeper linguistic analysis.
State-of-the-art detection systems address this complexity through supervised fine-tuning on expert-annotated corpora \citep{Spinde2021babe,Horych2024magpie}, but this approach is inherently limited: annotation is expensive, guidelines vary across projects, and models trained on one bias taxonomy generalize poorly to others.

We ask a different question: can \emph{structured multi-agent deliberation} serve as a competitive training-free alternative for media bias detection?
Large language models (LLMs) already possess the linguistic reasoning capacity to identify subtle bias cues \citep{Brown2020gpt3}, but single-agent approaches suffer from \emph{Degeneration-of-Thought}---self-reflection reinforcing initial judgments rather than revising them \citep{Liang2024mad}---and cannot simultaneously specialize in lexical analysis, evidence grounding, and framing detection.

Multi-agent debate (MAD) addresses this by having agents argue from distinct analytical angles \citep{Liang2024mad,Du2024debate}, though \citet{Cemri2025whyfail} caution that MAD systems suffer from recurrent failure modes and do not always outperform single-agent baselines. While MAD has been applied to factual claim verification, media \emph{bias} detection---concerning \emph{how} information is framed rather than whether it is true---remains largely unexplored in the debate paradigm.
Media bias detection is particularly well-suited to deliberative reasoning because bias often emerges from the interplay of multiple linguistic dimensions---lexical choice, evidential grounding, and structural framing---that no single analytical perspective can fully capture in isolation.

We propose \mabpd, a zero-shot pipeline that demonstrates our primary contribution: the Structured Argument Debate (SAD) framework. Rather than relying simply on multi-agent prompting, SAD replaces task-specific supervised decision boundaries with an explicit, principled deliberative structure.
Our key contributions are:

\begin{enumerate}
    \item A \textbf{Structured Argument Debate (SAD)} protocol that implements a domain-motivated asymmetric burden of proof---biased claims without grounded textual evidence carry zero weight---combined with role-weighted voting and evidence validation. Ablation confirms that this structured deliberation, not mere agent parallelism, is the primary performance driver ($+$9.3--10.6 F1).
    \item An \textbf{explicitly engineered multi-agent architecture} with functionally isolated agents (Bias, Evidence, Framing), an early-exit cascade for high-confidence neutral articles, and an asymmetric VerifierAgent (biased$\to$neutral only). A heterogeneous-LLM pilot (LLaMA 3.3 70B$+$Mixtral 8x7B$+$LLaMA 3.1 8B) changes F1 by only $\Delta{=}{-}0.1$~pp---the same order as run-to-run variation---indicating gains stem from the debate protocol, not model-specific artifacts.
    \item \textbf{Empirical evidence} that structured deliberation can match supervised training: on BABE (4{,}121 sentences, 23 topics), \mabpd achieves 83.4\% macro F1---within 0.7~pp of the supervised SOTA (MAGPIE, 84.1\%; \citealp{Horych2024magpie})---without any task-specific training, with cross-dataset transfer to SemEval 2019 HyperPartisan (75.0\% zero-shot accuracy, within 7.2~pp of supervised SOTA).
\end{enumerate}

\section{Related Work}
\label{sec:related}

\paragraph{Media Bias Detection.}
Computational approaches to media bias detection span lexicon-based
methods \citep{Recasens2013linguistic}, feature-engineered classifiers
\citep{Hube2019neural}, and transformer-based models fine-tuned on
annotated datasets \citep{Spinde2021babe,Wessel2023mbib}.
The BABE dataset \citep{Spinde2021babe} provides expert-level sentence
annotations across 23 topics and has become the standard benchmark;\footnote{The original BABE paper reports 3{,}700 sentences. We use the expanded 4{,}121-sentence version available on HuggingFace (\texttt{mediabiasgroup/BABE}), which includes additional annotation rounds. The 3{,}121/1{,}000 train/test split used in our experiments corresponds to this HuggingFace release.}
while MBIB \citep{Wessel2023mbib} unifies nine bias-detection subtasks.
\citet{Horych2024magpie} introduced MAGPIE, a multi-task approach pre-fine-tuned on 59 auxiliary bias tasks, achieving the current SOTA on BABE with 84.1\% macro F1.
Fine-tuned transformers achieve strong in-domain performance but depend on labeled data and generalize poorly across bias subtypes.

\paragraph{Multi-Agent Debate.}
\citet{Liang2024mad} introduced MAD to counter single-agent Degeneration-of-Thought, and \citet{Du2024debate} confirmed that debate improves factuality.
Subsequent work explored sparse topologies for cost reduction \citep{Li2024sparse}, confidence calibration \citep{Lin2025confmad}, systematic component ablation \citep{Becker2025mallm}, LLM-as-judge evaluation \citep{Chan2024chateval}, adaptive termination via stability detection \citep{Hu2025stabilitydebate}, and fake-news veracity \citep{Liu2025truedebate}. These target factual or preference judgements; \mabpd instead targets \emph{how} a claim is framed, and makes evidence grounding a scoring constraint rather than a prompt instruction.
\citet{Lin2025confmad} proposed ConfMAD, which weights agent votes by calibrated confidence; \mabpd extends this with a \emph{polarity-aware evidence factor} $f(\mathbf{e}, \text{claim})$ that imposes an asymmetric burden of proof: biased claims with zero textual evidence receive zero effective weight ($f{=}0$), whereas neutral claims retain partial weight ($f{=}0.6$) even without explicit evidence---a domain-motivated constraint absent from prior confidence-weighted schemes.
\citet{Han2025d2d} validated debate-for-detection for
\emph{misinformation} (D2D), showing that multi-agent deliberation
improves factual claim verification.
D2D assigns domain-specific profiles to each agent and employs
\emph{adversarial role-based specialization}: debater agents operate
from fixed opposing stances (affirmative vs.\ negative), a form of
structured diversity.
\mabpd differs in the \emph{type} of specialization and in three
architectural respects: (i)~\emph{analytical-dimension-based} agents
covering complementary perspectives (lexical bias, evidence quality,
structural framing) rather than adversarial stances;
(ii)~the SAD weighted scoring protocol with polarity-aware evidence
validation and asymmetric burden-of-proof (absent from D2D's uniform
agent weighting); and (iii)~an asymmetric post-consensus verifier.
Media \emph{bias}---concerning \emph{how} information is framed
rather than whether it is true---remains largely unexplored in the
debate paradigm; we address this gap.
Since neither reports on our benchmarks, we substitute two same-model controls
for a cross-paper comparison: a single-agent baseline and a BiasAgent-only
ablation (Tables~\ref{tab:ablation} and~\ref{tab:single_agent}). Both hold the
LLM fixed and vary only the architecture, attributing the improvement to SAD's
evidence grounding, debate and asymmetric verification rather than raw model
capability.

\begin{table}[htbp]
\centering
\small
\begin{tabular}{lll}
\toprule
\textbf{Framework} & \textbf{Evidence} & \textbf{Verifier Policy} \\
\midrule
MAD & None & Symmetric \\
D2D & Implicit & Symmetric \\
ConfMAD & None & Symmetric \\
\midrule
\textbf{\mabpd} & \textbf{Explicit} & \textbf{Asym (B$\to$N)} \\
\bottomrule
\end{tabular}
\caption{Architectural comparison of \mabpd vs existing MAD frameworks.}
\label{tab:novelty_comparison}
\end{table}

\paragraph{Criticism of MAD.}
\citet{Cemri2025whyfail} collected over 1{,}600 annotated execution
traces across seven multi-agent frameworks, developing the MAST
failure taxonomy (14 failure modes in three categories: system design
issues, inter-agent misalignment, and task verification failures)
from an initial analysis of 150 traces and validating it at scale.
Two findings bear directly on our design: verifier agents often perform superficial checks, and agents may disobey role specifications.
We address these by using a deterministic scoring agent, grounding debate arguments in verifiable textual spans, and employing an asymmetric verifier conservative about positive overrides.

\section{Methodology}
\label{sec:method}

\begin{figure*}[t]
\centering
\includegraphics[width=\textwidth]{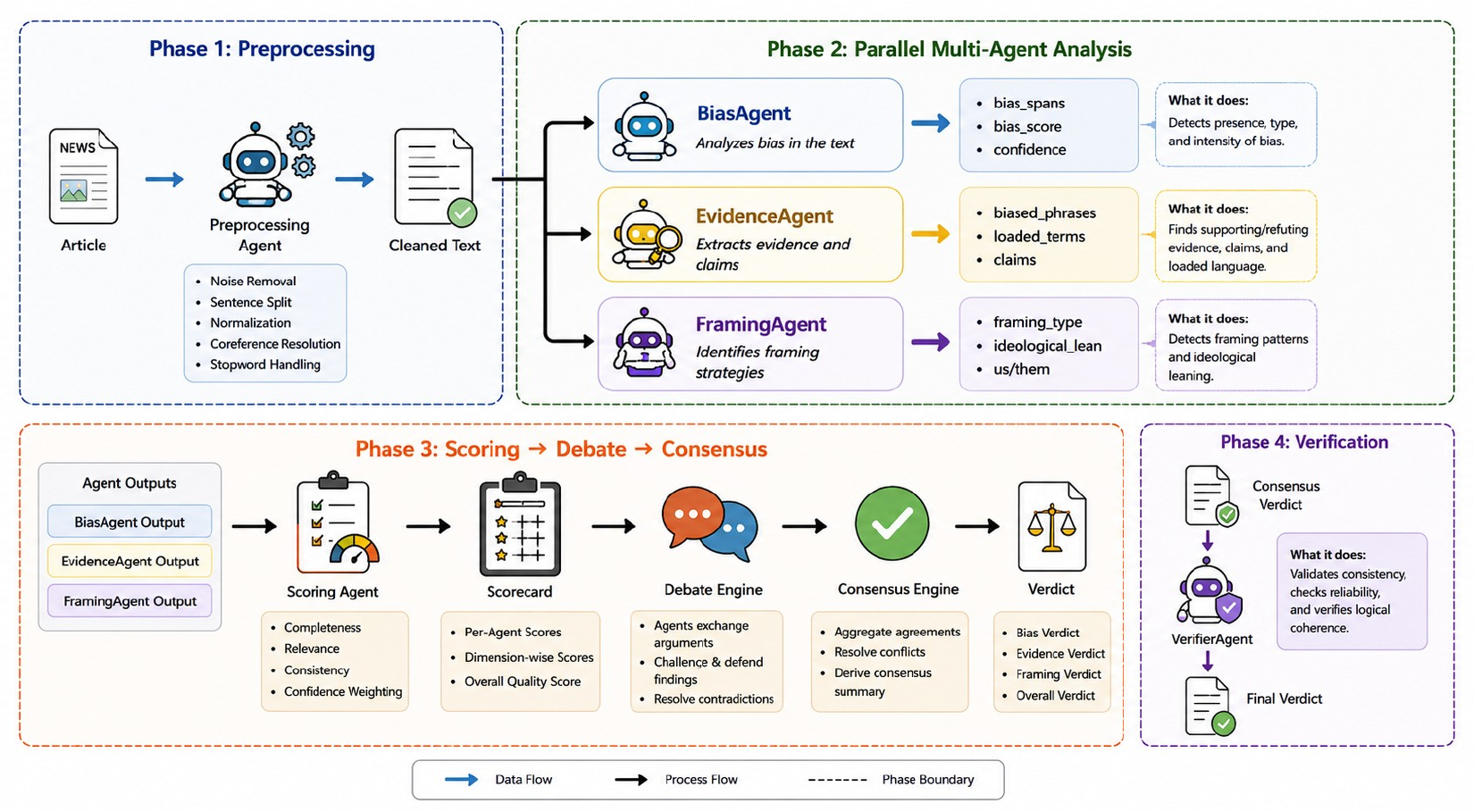}
\caption{Architecture overview of the \mabpd pipeline. Three
    functionally independent agents analyze text (BiasAgent first,
    then EvidenceAgent and FramingAgent without access to its output);
    their outputs are aggregated through a deterministic scoring agent
    and a three-path consensus engine (which may invoke the Structured
    Argument Debate protocol), and a VerifierAgent performs
    post-consensus validation.}
\label{fig:architecture}
\vspace{-3mm}
\end{figure*}

\mabpd is a pipeline organized into four phases: \emph{preprocessing}, \emph{multi-agent analysis}, \emph{debate and consensus}, and \emph{verification}.
Figure~\ref{fig:architecture} provides an overview.

\subsection{Phase 1: Preprocessing}
The \textbf{PreprocessingAgent} normalizes input text (encoding cleanup, whitespace normalization) and extracts metadata.
This step is deterministic and does not invoke the LLM for classification.

\subsection{Phase 2: Multi-Agent Analysis}
\label{sec:parallel}

Three specialized agents analyze the cleaned text independently, each providing a distinct analytical perspective.
The BiasAgent runs first; the EvidenceAgent and FramingAgent then run without access to its output, ensuring functional independence (execution order does not affect results).

\paragraph{BiasAgent.}
The primary bias detector.
It receives the article text along with bias category definitions from a structured knowledge base (10 categories; Appendix~\ref{sec:appendix-kb}) and classifies text using a 7-type taxonomy (Appendix~\ref{sec:appendix-taxonomy}).
Critically, these taxonomies and mapping tables were derived entirely from established media communication literature (e.g., \citealt{Wessel2023mbib,Fan2019babe}) and were strictly frozen prior to any exposure to the BABE dataset, ensuring no implicit prompt-optimization leaked into the zero-shot evaluation.
It returns a bias score $\in [0,1]$, a confidence score, and a list of bias spans with offsets and type labels, each mapped to one of four meta-categories via a 35-entry mapping table.
The BiasAgent supports \textbf{self-consistency} (SC) sampling \citep{Wang2022sc}: $k$ independent passes aggregated via majority vote.
As an optimization, when the BiasAgent returns a high-confidence neutral verdict ($c \geq 0.96$), the secondary agents are skipped via an early-exit cascade.

\paragraph{EvidenceAgent.}
A knowledge-augmented extraction agent that identifies span-level evidence of bias: biased phrases with quotes, loaded terms, and unsupported claims.
Loaded terms are matched against a hand-curated bias lexicon holding 793 terms at runtime, assembled from a released 451-entry, 26-category artifact and a larger set of in-module defaults; its construction, scoring, provenance and independence from the evaluation data are documented in Appendix~\ref{sec:appendix-lexicon}.
It validates evidence against a hand-curated protected-terms list of 234 entries across 11 topic domains---scientific, economic, medical, legal, security, immigration, and political-process vocabulary, together with competitive, cultural, and historical terminology---which must not be flagged as loaded language, reducing false positives; its curation is described in Appendix~\ref{sec:appendix-protected}.
Each extracted phrase includes an \texttt{evidence\_strength} rating (strong / moderate / weak).

\paragraph{FramingAgent.}
The FramingAgent applies 12 framing types, following the standard division between generic and issue-specific frames \citep{deVreese2005framing}. The seven generic frames comprise the episodic/thematic distinction \citep{Iyengar1991responsible} together with conflict, human-interest, attribution of responsibility, morality, and economic-consequences framing \citep{Semetko2000framing}. The five issue-specific frames extend established traditions: economic framing refines the economic-consequences frame; security framing follows the securitization literature \citep{Buzan1998security}; moral framing follows values-based accounts \citep{Lakoff2004elephant}; nationalist framing follows work on national-identity discourse \citep{Wodak2009national}; and public-health framing addresses population-level risk.
Returns the dominant framing type, ideological lean, us/them detection, and framing spans with explanations.

\paragraph{Evidence Store.}
A shared memory structure aggregates all agent outputs (bias spans, evidence phrases, framing spans) into a unified \textbf{EvidenceStore}, so that scoring and debate stages can draw on all agent findings jointly.

\subsection{Phase 3: Scoring, Debate, and Consensus}
\label{sec:consensus}

\paragraph{Deterministic Scoring.}
The \textbf{ScoringAgent} computes a multi-dimensional bias scorecard \emph{without} invoking an LLM (fully deterministic).
The overall bias score is:
\begin{equation}
\label{eq:scoring}
s_{\text{overall}} = 0.5 \cdot s_{\text{bias}} + 0.25 \cdot s_{\text{evidence}} + 0.25 \cdot s_{\text{framing}}
\end{equation}
where $s_{\text{bias}}$ incorporates the BiasAgent's bias score,
span count, and SC agreement factor; $s_{\text{evidence}}$ is derived
from the count of biased phrases, loaded terms, and unsupported
claims; and $s_{\text{framing}}$ reflects us/them detection,
ideological lean, and framing span count.
When SC agreement $\geq 80\%$, the bias component is boosted by $\times 1.15$ (if $\geq 80\%$ of runs detect bias) or attenuated by $\times 0.85$ (if $\geq 80\%$ agree neutral).

\paragraph{Three-Path Consensus Engine.}
The \textbf{ConsensusEngine} routes each article through one of three decision paths based on signal strength:

\begin{itemize}
    \item \textbf{Path A---Neutral Veto:} If the bias score is below a floor threshold (Tier-B: $s < 0.30$ with zero spans; Tier-A: $s < 0.40$ with SC neutral agreement $\geq 67\%$), the article is classified as neutral without debate.
    
    \item \textbf{Path B---Strong/Moderate Bias Fast-Path:} If $s \geq 0.85$ with $\geq 2$ high-confidence spans and secondary agent confirmation or abstention (or $s \geq 0.65$ with $\geq 1$ span and explicit secondary confirmation), the article is classified as biased.
    
    \item \textbf{Path C---Full Debate:} For ambiguous cases, the system invokes the Structured Argument Debate protocol (\S\ref{sec:sad}).
\end{itemize}

\paragraph{Structured Argument Debate (SAD).}
\label{sec:sad}
SAD structures the debate as a principled weighted agreement scoring framework.
Each of the three agents submits a structured argument:

\begin{equation}
\mathbf{a}_i = (\text{claim}_i, \; \mathbf{e}_i, \; c_i, \; \text{counter}_i)
\end{equation}
where $\text{claim}_i \in \{\text{biased}, \text{neutral}, \text{uncertain}\}$ is the agent's verdict, $\mathbf{e}_i$ is a list of supporting evidence spans ($|\mathbf{e}_i| \leq 5$), $c_i \in [0,1]$ is its self-reported confidence, and $\text{counter}_i$ is an anticipated counterargument.

The effective weight of each agent is:
\begin{equation}
\label{eq:weff}
w^{\text{eff}}_i = w_i \cdot c_i \cdot f(\mathbf{e}_i, \text{claim}_i)
\end{equation}
where $w_i$ is the architectural weight (BiasAgent: 0.50, EvidenceAgent: 0.25, FramingAgent: 0.25), $c_i$ is the adjusted confidence, and $f(\cdot)$ is a polarity-aware evidence factor (full table in Appendix~\ref{sec:appendix-weights}).
This design implements a \emph{burden of proof}: biased claims with zero evidence receive zero weight ($f=0$), while neutral claims without evidence retain partial weight ($f=0.6$).

The asymmetry in $f(\cdot)$ mirrors a courtroom burden of proof.
An agent claiming bias bears the obligation to present grounded textual evidence; without it, the claim carries zero effective weight---just as an unsupported legal assertion is dismissed regardless of the advocate's credentials.
Conversely, an agent claiming neutrality retains partial weight ($f{=}0.6$) even when it cites no explicit evidence, because the \emph{absence} of detectable bias markers is itself informative: a thorough search that finds no loaded language, unsupported claims, or selective framing constitutes meaningful evidence of neutrality.
Equation~\ref{eq:sad} formalizes this intuition.

The SAD score for each claim class $\mathcal{C}$ is:
\begin{equation}
\label{eq:sad}
\text{SAD}(\mathcal{C}) = \frac{\sum_{i: \text{claim}_i = \mathcal{C}} w^{\text{eff}}_i}{\sum_{i} w^{\text{eff}}_i}
\end{equation}

\begin{proposition}[Burden of Proof Safety]
\label{prop:sad_safety}
Let $\mathcal{A}$ be the set of agents. If agent $k \in \mathcal{A}$ claims $\mathcal{C} = \text{biased}$ but fails to provide valid textual evidence ($\mathbf{e}_k = \emptyset \implies f(\mathbf{e}_k, \text{biased}) = 0$), then $w^{\text{eff}}_k = 0$. Consequently, to achieve a biased consensus ($\text{SAD}(\text{biased}) \geq 0.50$), the combined effective weight of the remaining agents asserting bias must strictly equal or exceed the weight of agents asserting neutral. Thus, an ungrounded bias claim cannot unilaterally force a biased consensus.
\end{proposition}
\begin{proof}
Let $\mathcal{A}_B$ be the agents claiming ``biased''. The SAD score is $\text{SAD}(\text{biased}) = \frac{\sum_{i \in \mathcal{A}_B} w^{\text{eff}}_i}{\sum_{j \in \mathcal{A}} w^{\text{eff}}_j}$.
If $k \in \mathcal{A}_B$ has $f(\mathbf{e}_k, \text{biased}) = 0$, then $w^{\text{eff}}_k = 0$. For $\text{SAD}(\text{biased}) \geq 0.50$, we require $\sum_{i \in \mathcal{A}_B \setminus \{k\}} w^{\text{eff}}_i \geq \sum_{j \notin \mathcal{A}_B} w^{\text{eff}}_j$. If all agents in $\mathcal{A}_B$ lack evidence, the numerator is $0$, making a biased consensus mathematically impossible. Therefore, at least one agent must extract verifiable evidence to trigger a biased classification.
\end{proof}

The winning claim is
$\arg\max_{\mathcal{C}} \text{SAD}(\mathcal{C})$, subject to:
(i)~an evidence requirement (at least one corroborated bias span
across all biased-claiming agents), (ii)~a bias threshold
($\text{SAD}(\text{biased}) \geq 0.50$), and (iii)~tie-breaking by
evidence strength when scores are within 0.05.

Before scoring, SAD validates that cited evidence exists in the article text via substring matching (exact, fuzzy at 60\% overlap, partial at 30\%), preventing hallucinated evidence from influencing consensus.
Raw confidence scores are adjusted based on evidence quality and inter-agent agreement.
When SAD cannot resolve a conflict, the system falls back to majority vote; for non-trivial conflicts, an LLM-based \textbf{arbitrator} synthesizes agent arguments into a final decision.

\subsection{Phase 4: Verification}
\label{sec:verifier}

The \textbf{VerifierAgent} performs a three-part logic check: (1)~biased verdicts must have $\geq 1$ supporting span, (2)~biased verdicts with $s < 0.20$ are flagged as inconsistent, and (3)~neutral verdicts with $s \geq 0.80$ and $\geq 3$ spans are flagged as potential under-detection.
Critically, the Verifier operates under an \textbf{asymmetric override policy}: it may only change biased$\to$neutral (false-positive catch), never neutral$\to$biased.
For biased articles passing rule-based checks but with weaker primary evidence ($s < 0.65$ or fewer than 1 span), the Verifier additionally invokes the LLM to distinguish genuine author bias from factual reporting of partisan events; strong verdicts ($s \geq 0.65$ with $\geq 1$ span) bypass the LLM call.

\section{Experimental Setup}
\label{sec:experiments}

\subsection{Dataset}
We evaluate on the \textbf{BABE} (Bias Annotations By Experts) dataset, using the expanded 4{,}121-sentence HuggingFace release (\texttt{mediabiasgroup/BABE}; originally 3{,}700 sentences in \citealt{Spinde2021babe}), spanning 23 topics with a 55.8\% biased / 44.2\% neutral distribution (Table~\ref{tab:dataset}).
Since \mabpd is fully zero-shot---all thresholds are fixed architectural hyperparameters, not learned from BABE data---we report metrics on all 4{,}121 articles.

\begin{table}[t]
\centering
\small
\begin{tabular}{@{}lrrr@{}}
\toprule
\textbf{Statistic} & \textbf{Train} & \textbf{Test} & \textbf{Total} \\
\midrule
Sentences            & 3{,}121 & 1{,}000 & 4{,}121 \\
Biased              & 1{,}740 & 559     & 2{,}299 \\
Neutral             & 1{,}381 & 441     & 1{,}822 \\
Bias rate (\%)      & 55.8    & 55.9    & 55.8 \\
Mean word count     & 30.7    & 31.7    & 31.0 \\
Topics              & 23      & 23      & 23 \\
\bottomrule
\end{tabular}
\caption{BABE dataset statistics. Both splits maintain a similar bias rate ($\approx$56\%). ``Train'' = BABE official train partition (\texttt{ds[``train'']}); ``Test'' = held-out evaluation split (\texttt{ds[``test'']}). Since \mabpd uses no BABE data for training or threshold tuning, all 4{,}121 sentences are used for evaluation.}
\label{tab:dataset}
\end{table}

\subsection{Model Configuration}
\label{sec:model-config}
All agents use \textbf{LLaMA 3.3 70B Instruct} \citep{Dubey2024llama3} served via NVIDIA NIM API endpoints.
Using a single base model across all agents isolates the contribution of the multi-agent architecture and prompt specialization from model diversity effects.
The BiasAgent supports up to 5 self-consistency passes; in our primary evaluation we use SC${}=1$ for cost efficiency.

\subsection{Baselines}
We compare \mabpd against:
\begin{itemize}
    \item \textbf{BERT-base-uncased} \citep{Devlin2019bert} fine-tuned on BABE training data.
    \item \textbf{RoBERTa-base} \citep{Liu2019roberta} fine-tuned on BABE training data \citep{Spinde2021babe}.
    \item \textbf{MAGPIE (MTL:All)} \citep{Horych2024magpie}: RoBERTa pre-fine-tuned on 59 auxiliary bias tasks, the current published SOTA on BABE (84.1\% macro F1).
    \item \textbf{\mabpd (Single Agent)}: BiasAgent only, no evidence, framing, debate, or verifier.
    \item \textbf{\mabpd (No Debate)}: Full pipeline minus the debate module (consensus via rule-based paths only).
\end{itemize}

\subsection{Metrics}
We report accuracy, precision, recall, binary F1 (biased = positive class), and macro F1.
All runs are executed twice; we report the second run (run-to-run F1 difference $< 0.1$ pp; Table~\ref{tab:stability}).
Statistical significance is assessed via paired bootstrap \citep{Dror2018hitchhiker} with 10{,}000 resamples ($\alpha=0.05$).

\paragraph{Zero-shot scope.}
All decision thresholds (Table~\ref{tab:weights}) were fixed during system design, prior to any BABE evaluation.
Rather than relying on data-driven optimization, thresholds were derived explicitly from theoretical burden-of-proof constraints: the base classification boundary is mathematically fixed at 0.50 (``more likely biased than not''), and the evidence factors encode strict epistemic penalties derived from journalistic standards. All thresholds are defined as literal constants in the released source code, with no data-driven optimization step.

\section{Results}
\label{sec:results}

\subsection{Main Results}

Tables~\ref{tab:main_supervised} and~\ref{tab:main_zeroshot} present the main results under two evaluation regimes, separated to clarify the comparison context.

\begin{table}[t]
\centering
\small
\resizebox{\columnwidth}{!}{%
\begin{tabular}{@{}lccccc@{}}
\toprule
\textbf{System} & \textbf{Acc.} & \textbf{Prec.} & \textbf{Rec.} & \textbf{F1} & \textbf{MF1} \\
\midrule
BERT-base$^\ddagger$       & 74.1 & 71.4 & 76.8 & 74.0 & --- \\
RoBERTa-base$^\ddagger$    & 81.9 & 80.1 & \textbf{84.5} & 82.2 & --- \\
MAGPIE (MTL:All)$^\dagger$ & ---  & ---  & ---  & ---  & \textbf{84.1} \\
\mabpd (Full)              & \textbf{83.5} & \textbf{88.0} & 81.6 & \textbf{84.7} & 83.4 \\
\bottomrule
\end{tabular}}
\caption{Supervised Protocol --- Held-out Test Split (1{,}000 articles). Supervised baselines are fine-tuned on the 3{,}121-article BABE train partition; \mabpd uses no BABE data for training. Separating evaluation regimes ensures that \mabpd's zero-shot performance is compared against supervised models on identical test data.
    Bold marks the best value in each column, whichever system attains it. The F1 column is binary F1 and MF1 is macro F1; MAGPIE reports only macro F1, so it is listed in the MF1 column and left blank elsewhere. The regime-matched comparison is therefore macro-to-macro: \mabpd 83.4 vs.\ MAGPIE 84.1 ($-$0.7~pp).
    $^\dagger$84.1\% confirmed as macro F1 via source code inspection (\texttt{F1Score(average=``macro'')}, \texttt{head.py}).
    $^\ddagger$BERT and RoBERTa fine-tuned on the BABE train split (3{,}121 articles).}
\label{tab:main_supervised}
\end{table}

\begin{table}[t]
\centering
\small
\resizebox{\columnwidth}{!}{%
\begin{tabular}{@{}lccccc@{}}
\toprule
\textbf{System} & \textbf{Acc.} & \textbf{Prec.} & \textbf{Rec.} & \textbf{F1} & \textbf{MF1} \\
\midrule
\mabpd (Single)            & 71.1 & 90.1 & 54.0 & 67.6 & --- \\
\mabpd (No Debate)         & 78.1 & \textbf{93.1} & 65.6 & 76.9 & --- \\
\textbf{\mabpd (Full)}$^*$ & \textbf{84.7} & 87.2 & \textbf{85.2} & \textbf{86.2} & \textbf{84.6} \\
\bottomrule
\end{tabular}}
\caption{Zero-shot Protocol --- Full Dataset (4{,}121 articles). All three configurations use the same architectural thresholds with no training on BABE data. This regime isolates the contribution of each pipeline component under fully zero-shot conditions.
    Bold marks the best value in each column, whichever configuration attains it: the No-Debate arm has the highest precision, at a large cost in recall.
    $^*$Full system significantly outperforms No-Debate ($p{<}0.001$, paired bootstrap, 10K resamples).}
\label{tab:main_zeroshot}
\end{table}

On the full 4{,}121-article zero-shot evaluation, \mabpd achieves \textbf{84.7\% accuracy} and \textbf{86.2\% binary F1} (macro F1: 84.6\%).
Critically, on the \textbf{held-out 1{,}000-article test split} (regime-matched with supervised baselines), \mabpd achieves \textbf{84.7\% binary F1} (MF1: 83.4\%) and 83.5\% accuracy---\textbf{performing within 0.7~pp of the supervised SOTA (MAGPIE) without requiring any labeled training data}, and outperforming single-task RoBERTa by 2.5 F1 points.
A heterogeneous LLM pilot replacing secondary agents with Mixtral 8x7B and LLaMA 3.1 8B yields $\Delta$F1 = $-0.1$~pp (Appendix~\ref{sec:appendix-heterogeneous}), indicating that the observed gains are architectural rather than model-specific.
This confirms that the full-dataset result (86.2\%) is not an artifact of evaluating on articles overlapping with the supervised training split. Results are highly reproducible across two independent runs (run-to-run F1 difference = 0.06 pp; Table~\ref{tab:stability} in Appendix).

\subsection{Ablation Study}

Table~\ref{tab:ablation} presents the ablation results, showing the contribution of each pipeline component.

\begin{table}[t]
\centering
\small
\renewcommand{\arraystretch}{0.9}
\resizebox{\columnwidth}{!}{%
\begin{tabular}{@{}l@{\hspace{0.4em}}ccccr@{\hspace{1.2em}}ccccr@{}}
\toprule
 & \multicolumn{5}{c}{\textit{Full dataset (4{,}121)}} & \multicolumn{5}{c}{\textit{Test split (1{,}000)}} \\
\cmidrule(lr){2-6}\cmidrule(l){7-11}
\textbf{Config} & Acc & P & R & F1 & $\Delta$ & Acc & P & R & F1 & $\Delta$ \\
\midrule
Full System  & 84.7 & 87.2 & 85.2 & 86.2 & ---      & 83.5 & 88.0 & 81.6 & 84.7 & --- \\
$-$ Debate   & 78.1 & 93.1 & 65.6 & 76.9 & $-$9.3   & 76.0 & 93.2 & 61.5 & 74.1 & $-$10.6 \\
Single Agent & 71.1 & 90.1 & 54.0 & 67.6 & $-$18.6  & 69.7 & 91.0 & 50.8 & 65.2 & $-$19.5 \\
\midrule
Heterogeneous & \multicolumn{5}{c}{---} & 83.4 & 88.0 & 81.4 & 84.6 & $-$0.1 \\
\bottomrule
\end{tabular}}
\caption{Ablation study on both BABE splits. P = precision, R = recall; $\Delta$ = $\Delta$F1 vs.\ Full System. Every row was measured in the same serving environment (\S\ref{sec:model-config}), so all $\Delta$ values share a single reference point. The debate module contributes $-$9.3 F1 (full) and $-$10.6 (test split); multi-agent analysis overall contributes $-$18.6 (full) and $-$19.5 (test split) vs.\ single agent. Results are consistent across both evaluation regimes.}
\label{tab:ablation}
\end{table}

\begin{figure}[t]
\centering
\includegraphics[width=\columnwidth]{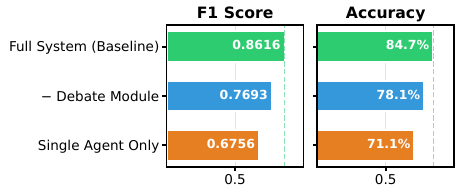}
\caption{F1 and accuracy comparison across ablation scenarios. The full system substantially outperforms both reduced configurations, with the gap driven primarily by recall improvements from the debate module.}
\label{fig:ablation-perf}
\end{figure}

Figure~\ref{fig:ablation-perf} visualizes the performance differences across configurations.

Removing debate drops F1 by 9.3 points on the full dataset (86.2 $\to$ 76.9) and by 10.6 points on the test split (84.7 $\to$ 74.1), primarily through \textbf{recall}: without debate, recall falls from 85.2\% to 65.6\% on the full dataset ($-$19.6 pp) and from 81.6\% to 61.5\% on the test split ($-$20.1 pp), while precision remains high ($>$93\% in both regimes).
This confirms that rule-based consensus paths (A and B) are conservative and miss subtle bias cases requiring full SAD deliberation.
The single-agent configuration has high precision ($>$90\%) but catastrophically low recall ($<$55\%); adding the secondary agents without debate recovers 8.9--9.3 F1 points, and adding the debate module on top of them recovers a further 9.3--10.6.

Crucially, the debate effect is \emph{larger} on the harder test split ($-$10.6 vs.\ $-$9.3), indicating that structured deliberation is most valuable on the in-domain, expert-curated articles that supervised models are trained on. Additionally, replacing the EvidenceAgent and FramingAgent with Mixtral 8x7B and LLaMA 3.1 8B changes performance by only $\Delta$F1 = $-0.1$~pp, providing evidence that the architecture is robust against single-model correlated errors (detailed in Appendix~\ref{sec:appendix-heterogeneous}).

The ablations in Table~\ref{tab:ablation} vary the debate module and the presence of the secondary agents as a group; we do not report an ablation of each secondary agent in isolation (see Limitations). The heterogeneous pilot provides indirect evidence on the same question: substituting substantially weaker models for both secondary agents moves F1 by only $-$0.1~pp (Appendix~\ref{sec:appendix-heterogeneous}), indicating that no single secondary agent carries the architecture's advantage, whereas removing the mechanism that adjudicates between them costs an order of magnitude more.

\subsection{Confusion Matrix Analysis}

Table~\ref{tab:confusion_appendix} (in Appendix~\ref{sec:appendix-cm}) shows the confusion matrices for all three system configurations.
The full system shows balanced errors (288 FP, 341 FN; ratio 0.84:1), unlike the single-agent ablation where FN errors dominate (1{,}057 vs.\ 136, ratio 7.8:1).
The full pipeline recovers 716 previously-missed biased articles while adding only 152 false positives---a favorable 4.7:1 trade-off.

\subsection{Evaluation Regime}
\label{sec:eval-regime}

We evaluate \mabpd on both the full BABE dataset (4{,}121 articles) and the held-out test split (1{,}000 articles). While zero-shot systems lack a train/test distinction, reporting on the test split ensures fair comparison with supervised baselines \citep{Spinde2021babe}. On this split, \mabpd achieves \textbf{84.7\% binary F1} and \textbf{83.4\% macro F1}, remaining highly competitive with the supervised SOTA (MAGPIE, 84.1\% MF1) despite having no parameter tuning.

\subsection{Single-Agent LLM Baseline Comparison}
\label{sec:baseline-comparison}

To validate whether the performance gains stem from the multi-agent architecture rather than the underlying LLM's raw capability, we compared \mabpd against a strong single-agent baseline (Qwen-3.5 122B-A10B). As shown in Table~\ref{tab:single_agent}, \mabpd significantly outperforms the single-agent setup, indicating that Structured Argument Debate provides capabilities beyond standard prompting.

\begin{table}[htbp]
\centering
\small
\resizebox{\columnwidth}{!}{%
\begin{tabular}{@{}lccccc@{}}
\toprule
\textbf{Model Setup} & \textbf{Acc} & \textbf{Prec} & \textbf{Rec} & \textbf{Bin F1} & \textbf{Mac F1} \\
\midrule
Single-Agent (Qwen-3.5) & 80.5\% & 93.5\% & 69.9\% & 80.0\% & 80.5\% \\
\textbf{\mabpd (Ours)} & \textbf{83.5\%} & \textbf{88.0\%} & \textbf{81.6\%} & \textbf{84.7\%} & \textbf{83.4\%} \\
\bottomrule
\end{tabular}}
\caption{Single-Agent LLM Baseline Comparison on the BABE test split. \mabpd demonstrates substantial improvements over a strong external single-agent baseline (Qwen-3.5 122B-A10B MoE via NVIDIA NIM).}
\label{tab:single_agent}
\end{table}

\subsection{Error Analysis}

The full system produces 288 false positives and 341 false negatives on the full dataset (4,121 articles).
False positives predominantly involve \emph{factual political reporting}---the pipeline detects strong framing signals in articles reporting on partisan events even when the author's own voice is neutral---and \emph{emotionally valenced neutral content} (disaster reports, crime coverage) where loaded vocabulary is incidental rather than authorial.

To quantify false negative patterns, we manually categorized 100 randomly sampled false negative articles. The distribution confirms \emph{omission bias} as the dominant failure mode (46\%), followed by \emph{source-selection bias} (32\%), and \emph{subtle epistemological framing} (22\%). Because omission bias requires external knowledge about what is \emph{missing}, LLMs relying solely on parametric memory struggle to detect it; retrieval-augmented generation is a promising direction to address this structural limitation.

\subsection{Qualitative Examples}
\label{sec:qualitative}

We present a qualitative example illustrating how structured deliberation identifies implicit bias where a single-agent baseline fails. Two additional examples spanning left-leaning and right-leaning framing bias are provided in Appendix~\ref{sec:appendix-qualitative-extra}.

\begin{table}[htbp]
\centering
\scriptsize
\begin{tabularx}{\columnwidth}{@{}l X@{}}
\toprule
\textbf{Component} & \textbf{Output} \\
\midrule
\textbf{Snippet (GT: Biased)} & ``The concept of changing one's biological sex is, of course, nonsense, as sex is determined by unalterable chromosomes.'' \\
\textbf{Single-Agent Baseline} & \textbf{Neutral} (Conf: 0.501. Misses loaded implication.) \\
\midrule
\textbf{BiasAgent Evidence} & Extracted loaded language (``nonsense'') and ideological framing. \\
\textbf{EvidenceAgent Evidence} & Extracted subjective phrase (``of course, nonsense''). \\
\textbf{FramingAgent Evidence} & Predicted \textbf{Neutral} (Failed to detect structural bias). \\
\midrule
\textbf{SAD Debate Reasoning} & The explicit loaded language and unsupported claims provided stronger evidence, overriding the neutral perspective. \\
\textbf{Final \mabpd Verdict} & \textbf{Biased} (Consensus Confidence: 0.856) \\
\bottomrule
\end{tabularx}
\caption{Example~1: Epistemological Framing Bias. \mabpd captures subtle epistemological bias missed by a strong single-agent baseline.}
\label{tab:qualitative}
\end{table}

\subsection{Human Evaluation}
\label{sec:human-eval}

To validate real-world utility, we conducted a human evaluation on 150 stratified articles with three volunteer annotators (PhD scholars with exposure to NLP and media
studies, fluent in English), familiarized with the BABE taxonomy.
Annotators evaluated each article independently, without access to model predictions or each other's labels.
\mabpd achieved 89.3\% accuracy against the human majority vote (Fleiss' $\kappa = 0.66$), with 92\% agreement on biased articles and 87\% on neutral articles---well above the $\kappa < 0.10$ pitfalls common in crowd-sourced bias annotation on subjective political texts.

\subsection{Cost and Efficiency Analysis}
\label{sec:efficiency}

The full SAD system requires $\sim$6.8$\times$ the tokens of a single-agent pass ($\sim$8,200 vs.\ $\sim$1,200), but the absolute cost remains economical (\$0.0065 vs.\ \$0.0010/article). Per-article \emph{end-to-end} latency---one article in isolation---scales sub-linearly with the agent count (16.5s full vs.\ 4.5s no-debate vs.\ 1.6s single-agent); under concurrent dispatch the \emph{amortized} figure is far lower (Appendix~\ref{sec:appendix-efficiency}). The early-exit cascade is designed to skip secondary agents for high-confidence neutral articles (confidence $\geq 0.96$); on the bias-heavy BABE corpus, this condition rarely triggered, but the mechanism targets neutral-skewed deployment corpora where cost savings are most impactful.

\subsection{Analysis}
\label{sec:analysis}

We structure our analysis around three research questions.

\textbf{RQ1: Does structured debate contribute beyond agent parallelism?}
The 9.3-point F1 gap between the full system and no-debate ablation demonstrates that ambiguous cases require full SAD deliberation. The burden-of-proof mechanism prevents agents from asserting bias without grounding, while the VerifierAgent's asymmetric override catches false positives. Per-article error analysis confirms this is overwhelmingly a recall mechanism: of 718 articles corrected vs.\ single-agent, 716 (99.7\%) are rescued false negatives at the cost of 152 additional false positives (net $+$564 correct).

\textbf{RQ2: Is the decision threshold principled?}
Sweeping the SAD threshold reveals that our theoretical 0.50 baseline achieves the highest macro F1 (83.4\%); across the practical range (0.35--0.70), all thresholds remain within 1.5~pp of the optimum (full sweep in Appendix~\ref{sec:appendix-threshold}).

\textbf{RQ3: Does \mabpd generalize?}
On the SemEval 2019 HyperPartisan News dataset \citep{Kiesel2019hyperpartisan} (644 articles, no overlap with BABE), \mabpd achieves \textbf{75.0\% zero-shot accuracy}, within 7.2~pp of supervised SOTA. The recall-heavy pattern persists (recall 88.8\%, precision 61.4\%), confirming cross-corpus transfer.

\section{Conclusion}
\label{sec:conclusion}

We presented \mabpd, a multi-agent pipeline without task-specific fine-tuning whose core contribution---the Structured Argument Debate (SAD) protocol---replaces task-specific supervised decision boundaries with an explicit asymmetric burden of proof, role-weighted voting, and evidence grounding.
Ablation isolates structured deliberation as the primary driver: removing debate costs up to 10.6 F1 points, driven by recall collapse ($-$20.1~pp).
On the BABE held-out test split, \mabpd achieves 83.4\% macro F1---within 0.7~pp of supervised SOTA \citep{Horych2024magpie}---without any task-specific training; cross-dataset transfer to HyperPartisan yields 75.0\% zero-shot accuracy (within 7.2~pp of supervised SOTA; \citealt{Kiesel2019hyperpartisan}).
At an amortized 2.0~s/article under concurrent dispatch, the system is practical for corpus-scale deployment.
Future work includes heterogeneous base models \citep{Lin2025llmbias}, sparse debate topologies \citep{Li2024sparse}, multilingual extension, and failure trace annotation \citep{Cemri2025whyfail}.

\section*{Limitations}
\label{sec:limitations}

\paragraph{Single model and agent independence.}
All agents in the main experiments use the same LLM (LLaMA 3.3 70B), isolating architectural contributions but sharing latent biases that may create correlated errors \citep{Lin2025llmbias}.
Because agents differ only in prompt specialization---not in model weights---their disagreements are driven by prompt engineering rather than genuine epistemic diversity.
We quantify this concern in Appendix~\ref{sec:appendix-independence} and further mitigate it via the heterogeneous LLM pilot study (Appendix~\ref{sec:appendix-heterogeneous}), which confirms the architecture is model-agnostic.

\paragraph{Scope, generalizability, and comparability.}
We evaluate exclusively on BABE (English only, 23 topics, $\approx$56\% bias rate) for our primary metrics.
Although our thresholds were not tuned on BABE data, they were designed for a binary bias-detection task with balanced class distributions.
While cross-dataset evaluation on the SemEval 2019 HyperPartisan corpus (\S\ref{sec:analysis}) confirms broad transferability, we observe a significant precision-recall imbalance on this corpus (88.8\% recall vs. 61.4\% precision). This suggests that zero-shot thresholds derived from general media bias definitions may over-trigger on the highly stylized, opinionated language common in HyperPartisan texts, requiring future adaptation.
Two properties of that run should be stated. Articles were truncated to their first 600 words before dispatch (\texttt{prepare\_hyperpartisan.py}), affecting 206 of the 644, so this is not a full-text result; and the run covers 644 of the 645 \texttt{byarticle} articles. Accuracy is higher on the truncated subset (80.6\% vs.\ 72.4\%), so truncation does not appear to drive the precision loss, but bias residing later in a long article is invisible by construction.

Our comparison with MAGPIE is subject to one key caveat: examination of the MAGPIE source code (\texttt{head.py}) confirms that 84.1\% is \textbf{macro F1} (\texttt{F1Score(average=``macro'')}).
The regime-matched comparison is therefore \mabpd 83.4\% macro F1 vs.\ MAGPIE 84.1\% macro F1 ($-$0.7~pp), zero-shot vs.\ supervised on the 1{,}000-article held-out test split.
Supervised baselines train on the 3{,}121-article BABE train partition; \mabpd uses no BABE data for training or threshold tuning (\S\ref{sec:eval-regime}).

A direct head-to-head against D2D \citep{Han2025d2d} and ConfMAD \citep{Lin2025confmad} is likewise unavailable: neither targets media bias, and neither reports on BABE or HyperPartisan, so no published number exists to compare against.
We therefore isolate the architecture with same-model controls (Tables~\ref{tab:ablation} and~\ref{tab:single_agent}), which hold the LLM fixed and vary only the pipeline, rather than with a cross-paper comparison.
A stricter test would hold one set of agent outputs fixed and vary only the aggregation rule---majority vote, unweighted debate, and confidence-weighted voting with the evidence factor removed---thereby separating our aggregation from these schemes directly.
We do not report it: the released artifacts retain only final verdicts and confidences rather than per-agent scores, and the serving environment in which the reported runs were measured is no longer available (see \emph{Serving environment and reproducibility} below), so the agent outputs cannot be regenerated under the original conditions.
We regard this as the most informative next comparison for future work.

\paragraph{Serving environment and reproducibility.}
\label{par:serving}
All results reported in this paper were obtained through a single serving environment: LLaMA 3.3 70B Instruct accessed via the NVIDIA NIM API (\S\ref{sec:model-config}).
That model endpoint now carries a provider deprecation notice and is no longer reachable, so we were unable to re-run the reported configurations under the environment in which they were measured.
Two consequences follow. First, because every number comes from one environment, we cannot separate effects attributable to the model weights from effects attributable to the serving stack---quantization profile, server-side sampling defaults, and model revision are not observable to an API client and were not recorded at run time.
Second, absolute values obtained through a different provider, a different quantization profile, or a later model revision may differ from those reported here: \citet{Yuan2025nondeterminism} show that changing evaluation batch size, GPU count, or GPU revision alone can alter generated text and shift benchmark accuracy even under greedy decoding, because floating-point arithmetic is non-associative at limited numerical precision.
We would expect the relative ordering of components to be more robust than their exact magnitudes, but we cannot demonstrate this from a single environment.
We also note a concrete improvement we did not adopt: our client does not set a server-side random seed, and doing so would make future runs bit-reproducible at negligible cost.
Readers reproducing this work should therefore treat the reported figures as specific to the stated environment and expect variation elsewhere.

\paragraph{Per-agent ablation.}
Table~\ref{tab:ablation} isolates the debate module and the secondary agents as a group, but does not report an ablation of the EvidenceAgent and the FramingAgent individually. We do not include those two configurations here: the environment in which the reported ablations were measured is no longer available, and a per-agent ablation measured elsewhere would not be comparable row-by-row with the rest of Table~\ref{tab:ablation}. We regard a per-agent ablation, measured end-to-end in a single controlled environment alongside the configurations already reported, as the natural next step.

\paragraph{Known MAS failure modes.}
Despite mitigations, our system remains susceptible to the failure classes cataloged by \citet{Cemri2025whyfail}, and in particular to echo-chamber convergence and sycophantic consensus at the verification stage. We provide a detailed trace annotation study in Appendix~\ref{sec:appendix-failure-modes}.

\paragraph{Deployment guardrails.}
SAD already validates that every cited span occurs verbatim in the article, so
fabricated evidence cannot sway a verdict. For deployment we recommend two
further lightweight, post-hoc checks aimed at the sycophancy failure mode. A
\emph{disagreement check} flags an article for human review when the agents
disagree strongly yet the Verifier converges with high confidence. A
\emph{counter-argument step} extends the Verifier prompt to require the strongest
case for \textsc{neutral} before it decides. Both are heuristics over existing
outputs and require no retraining.

\paragraph{Extensions not evaluated here.}
Two extensions identified during review are outside the scope of the reported
results: granularity-aware dynamic weighting---scaling the FramingAgent's weight
with input length---and evaluation on the broader MBIB collection. Neither is
evaluated in this paper, and no claim here should be read as covering them.

\section*{Ethical Considerations}

Bias detection systems risk suppressing legitimate perspectives or mislabeling factual reporting.
We mitigate this through the asymmetric verifier (conservative about bias labels), expert-annotated evaluation data, and full code release.
The underlying LLM may harbor political biases \citep{Lin2025llmbias}; we emphasize that our system is a \emph{tool for awareness}, not an automated censorship mechanism, and recommend human review before consequential decisions.

\mabpd is not designed to replace professional journalists, editors or media
analysts, nor to act as an automated arbiter of truth. It is positioned as a
first-pass awareness aid that surfaces potentially biased passages for human
attention and supports media literacy. Its output is an early-warning signal
rather than a verdict, agent consensus should not be treated as ground truth, and
a human should remain in the loop for any consequential decision---a caution
reinforced by the failure modes documented in
Appendix~\ref{sec:appendix-failure-modes}, including sycophancy and reduced
out-of-domain precision.

\section*{Acknowledgements}

We used LLM assistants (ChatGPT, Claude) and an AI coding assistant (Kiro) in the preparation of this work.
AI tools were used to review and revise prose written by the authors, including improving and paraphrasing selected passages (e.g., \S\ref{sec:efficiency}, the Abstract, and the Conclusion) and editing for clarity and LaTeX formatting throughout; no part of the manuscript was generated by an AI system from instructions.
The coding assistant was used for code scaffolding, evaluation script generation, and analysis utilities.
All AI-revised content was reviewed, verified, and approved by the authors, who take full responsibility for the accuracy of all claims.
The \mabpd system's architecture, experimental design, implementation decisions, and result interpretation are the authors' own intellectual contributions.

\bibliography{custom}

\appendix

\FloatBarrier

\begin{table*}[t]
\centering
\small
\begin{tabularx}{\textwidth}{@{}lrX@{}}
\toprule
\textbf{Parameter} & \textbf{Value} & \textbf{Justification} \\
\midrule
BiasAgent weight ($w_1$) & 0.50 & 2:1:1 agent weighting: the BiasAgent is the primary detector, carrying twice the weight of the supporting agents. \\
EvidenceAgent weight ($w_2$) & 0.25 & Supporting agent under the 2:1:1 weighting. \\
FramingAgent weight ($w_3$) & 0.25 & Supporting agent under the 2:1:1 weighting. \\
\midrule
Evidence factor $f$ (biased, $|\mathbf{e}|=0$) & 0.00 & Burden-of-proof axiom: a bias claim with no grounded textual evidence carries zero weight. \\
Tier-B neutral veto ceiling & 0.30 & Deliberately conservative; demands stronger evidence before flagging bias. \\
Tier-A neutral veto ceiling & 0.40 & Deliberately conservative; demands stronger evidence before flagging bias. \\
SC neutral agreement floor & 67\% & Set conservatively; a clear majority of self-consistency passes must agree. \\
Strong-bias score floor & 0.85 & Conservative margin in the same direction. \\
Moderate-bias score floor & 0.65 & Conservative margin in the same direction. \\
HC span confidence floor & 0.60 & Conservative margin in the same direction. \\
SAD bias threshold & 0.50 & Decision boundary at ``more likely biased than not''. \\
\midrule
\multicolumn{3}{@{}p{\textwidth}@{}}{\emph{Sensitivity:} across 0.35--0.70 macro F1 stays within 1.5 points of the optimum (Appendix~\ref{sec:appendix-threshold}, Table~\ref{tab:threshold-sweep}).} \\
\bottomrule
\end{tabularx}
\caption{Architectural weights and decision thresholds, with the design rationale for each value.}
\label{tab:weights}
\end{table*}

\section{Agent Weights and Thresholds}
\label{sec:appendix-weights}

Table~\ref{tab:weights} lists the architectural weights and key thresholds used in the SAD protocol and consensus engine.
The polarity-aware evidence factor $f(\mathbf{e}, \text{claim})$ used in Eq.~\ref{eq:weff} is:
\begin{equation}
\label{eq:efactor}
f(\mathbf{e}, \text{claim}) = \begin{cases}
0.6 & \text{if claim = neutral}, |\mathbf{e}|=0 \\
1.0 & \text{if claim = neutral}, |\mathbf{e}|>0 \\
0.0 & \text{if claim = biased}, |\mathbf{e}|=0 \\
0.6 & \text{if claim = biased}, |\mathbf{e}|=1 \\
0.85 & \text{if claim = biased}, |\mathbf{e}|=2 \\
1.0 & \text{if claim = biased}, |\mathbf{e}| \geq 3
\end{cases}
\end{equation}

\section{Bias Taxonomy}
\label{sec:appendix-taxonomy}

The BiasAgent uses a 7-type bias taxonomy:
\begin{enumerate}
    \item \textbf{Loaded language}: emotionally charged words designed to influence.
    \item \textbf{Framing bias}: selective emphasis that shapes interpretation.
    \item \textbf{Selection bias}: cherry-picking facts or sources.
    \item \textbf{Emotional language}: appeals to emotion rather than reason.
    \item \textbf{Political labeling}: ideological labels used pejoratively.
    \item \textbf{Exaggeration}: overstatement beyond what evidence supports.
    \item \textbf{Omission bias}: deliberate omission of relevant counter-evidence.
\end{enumerate}

These are mapped to four bias meta-categories---biased language, framing bias, unsupported claims, and selective presentation---via a 35-entry mapping table.

\section{Knowledge Base Categories}
\label{sec:appendix-kb}

Table~\ref{tab:kb} lists the 10 bias categories defined in the structured knowledge base (\texttt{bias\_definitions.json}) that is injected into agent prompts.

\begin{table}[htbp]
\centering
\scriptsize
\begin{tabularx}{\columnwidth}{@{}llX@{}}
\toprule
\textbf{Category} & \textbf{Sev.} & \textbf{Definition (abbreviated)} \\
\midrule
Loaded language        & High   & Words/phrases with strong emotional connotations used to influence perception. \\
False balance          & Med.   & Presenting two sides as equally valid when evidence supports one. \\
Cherry-picking         & High   & Selective use of facts/data supporting a predetermined narrative. \\
Political framing      & High   & Presenting information favoring a particular political position. \\
Dog whistle            & High   & Language appearing neutral but carrying coded connotations. \\
Scapegoating           & High   & Blaming a specific group for complex problems without evidence. \\
Framing bias           & Med.   & Presentation style that shapes audience interpretation. \\
Unsupported claim      & Med.   & Assertions presented as fact without evidence or attribution. \\
Biased language        & High   & Language reinforcing prejudice, stereotypes, or assumptions. \\
Selective presentation & Med.   & Including/excluding information to support a particular narrative. \\
\bottomrule
\end{tabularx}
\caption{Knowledge base bias categories injected into agent prompts.}
\label{tab:kb}
\end{table}

\section{Construction of the Protected-Terms List}
\label{sec:appendix-protected}

The protected-terms list is a hand-curated set of 234 entries spanning 11 topic
domains: climate science (30), economics (38), public health (23), democratic
process (32), immigration (17), legal (18), military and security (14), science
and technology (14), sport (21), culture and media (11), and academic and
historical vocabulary (16). It holds ordinary, editorially neutral
terminology---for example \emph{interest rate}, \emph{vaccination}, \emph{polling
day}---that language models tend to mis-flag as loaded simply because it
co-occurs with politically charged topics. Terms were compiled from established
media-bias literature together with authoritative domain references, including
IPCC terminology for climate, WHO and CDC usage for public health, and
central-bank glossaries for economics, following the same literature-grounded
design as the bias taxonomy and frozen prior to any exposure to BABE. One term,
\emph{defence}, is registered under two domains (legal and military/security),
so the list contains 233 distinct strings. The list is emitted by the released
deterministic builder \texttt{knowledge/build\_lexicon.py}. At inference the
EvidenceAgent discards any protected term the model proposes as loaded language,
reducing false positives on domain vocabulary. No BABE article, label or split
informed its construction, preserving the zero-shot setting.

\section{Construction of the Bias Lexicon}
\label{sec:appendix-lexicon}

\paragraph{Structure.}
The lexicon comprises 451 entries across 26 categories (449 unique surface
forms; two terms appear under two categories). Each entry is a 4-tuple---term,
bias score, neutral alternative, residual flag---maintained in
\texttt{knowledge/build\_lexicon.py}, from which all released JSON artifacts are
generated. The category scheme is our own operationalisation, informed by the
bias dimensions surveyed in \citet{Hamborg2019review} and the task groupings of
\citet{Wessel2023mbib}, and by the word- and sentence-level annotations of
\citet{Spinde2021babe}; category names and boundaries are ours and do not
correspond one-to-one to categories in those works. The lexicon covers English,
in the domain of US political and news discourse.

\paragraph{Candidate identification.}
Terms were written by the authors from the category definitions, informed by the
media-bias literature. Established coded terms and political epithets were
included where their usage is documented in general reference works; the
remainder are compositional phrases and rhetorical constructions characteristic
of their category. No term list was copied from an existing lexicon resource.

\paragraph{Scoring.}
Scores follow a four-anchor ordinal scale documented in the builder
source---0.60 mild, 0.70 moderate, 0.80 strong, 0.90 and above very strong. 84
entries sit exactly on an anchor and 16 above 0.90; the remaining values order
terms within a band rather than expressing calibrated magnitudes. Band placement
reflects how far a term's evaluative force is independent of context rather than
how offensive the term is in isolation: terms that are almost exclusively
evaluative sit in the upper bands, while 16 terms whose predominant usage is
neutral sit below the mild anchor with a rationale recorded inline in the source
(for example \emph{alleged} at 0.40, ``neutral legal/journalistic usage'';
\emph{champion} at 0.50, ``sports/advocacy neutral''). The same criterion
motivates the protected-terms list. Within a band, intensified variants generally
score above their base form (\emph{dangerous} 0.50 / \emph{gravely dangerous}
0.85; \emph{incompetent} 0.87 / \emph{gross incompetence} 0.92). Detection
selects the highest-scoring matching term per sentence under a 0.45 confidence
floor, so entries below that floor are inactive.

Scores were assigned by the authors from the category definitions. They were not
obtained by multi-rater annotation and we report no inter-annotator agreement
statistic. We note that agreement on word-level bias is low even under trained
expert protocols: \citet{Spinde2021reliable} report Krippendorff's
$\alpha = 0.144$ for crowdsourced and $0.419$ for expert labels on this task,
and \citet{Lim2020annotating} likewise find bias-word judgements substantially
subjective.

\paragraph{Sensitivity of the scores.}
Because lexical bias measures are known to be sensitive to word-list
construction \citep{Antoniak2021badseeds}, we quantify what the score values
contribute. Re-running lexicon detection over the 1{,}173 BABE records
containing at least one lexicon term under progressively coarser scales:
collapsing 19 distinct values to 8 leaves 99.7\% of selected evidence unchanged;
four anchors leave 96.7\% unchanged; removing score information entirely leaves
90.7\% unchanged. The scores therefore act as ordinal tie-breakers among
co-occurring terms, and their fine-grained differences are consequential in under
1\% of cases. This analysis is reproducible from the released files without model
access.

\paragraph{Independence from the evaluation data.}
No BABE article, label or split informed term membership or scoring. Three checks
support this: 336 of 449 terms (74.8\%) occur in none of the 4{,}121 BABE
records; the correlation between assigned score and the bias rate of containing
records is $r = +0.271$ ($n = 48$, critical $r = 0.285$, not significant); and
BABE's only numeric field is a binary label, so no graded score could have been
read from it.

\paragraph{Scope of the released artifact.}
The runtime lexicon is assembled from two sources, and we state the split
explicitly because it bears directly on reproducibility. The EvidenceAgent module
declares an in-module dictionary of 560 terms; at import,
\texttt{knowledge/bias\_lexicon.json} (449 terms) is merged \emph{on top} of it.
Of the released terms, 218 are already present in the in-module dictionary and
their JSON score overrides the in-module value, while 231 are new and are added.
The remaining \textbf{342 in-module terms are unique to the source code}: they are
not contained in any released JSON artifact, yet they are active in every run and
constitute 43\% of the 793-term runtime lexicon.\footnote{The merged lexicon holds
793 keys but 791 case-insensitively distinct surface forms: two in-module entries
(\emph{real Americans}, \emph{real Australians}) are stored alongside their
lower-cased counterparts supplied by the JSON, so each is counted twice. Matching
is case-insensitive, so this affects the reported size only, not detection.}
We emphasise that these 342 terms are \emph{not} a fallback. The merge is
unconditional and additive, so they participate in detection whenever the JSON
loads successfully; only their scores are unaffected by the released file. A
reader who reproduces from \texttt{knowledge/*.json} alone therefore obtains 449
terms rather than 793 and should expect different span-level output. The
in-module dictionary is listed in
\texttt{pipeline/agents/evidence\_agent.py}, and consolidating it into the
released builder is planned future work.

\paragraph{Provenance and status of the in-module terms.}
The 342 code-only terms have the same origin as the released ones: they were
written by the authors from the category definitions and the media-bias
literature, and no term list was copied from any existing lexical resource.
Established coded terms and political epithets were included where their usage is
documented in general reference works; the remainder are compositional phrases
characteristic of their grouping. We searched for a prior collection matching this
term set and found none: no published lexicon we are aware of shares the schema,
and in particular the \emph{neutral alternative} field appears in none of HurtLex,
MPQA, the NRC lexicons, the General Inquirer, or existing media-bias word lists.
Estimated incidental overlap with the nearest comparable resource, a documented
dog-whistle glossary, is under 2\%, consistent with independent construction over
shared subject matter. Because this content is authored lexicon material that
happens to reside in a \texttt{.py} file, the repository licences it as data as
well as code: the in-module defaults are offered under CC-BY-4.0 in addition to
the repository's MIT licence, so a reader may extract and reuse them on the same
terms as the released JSON. Roughly 45 entries across the JSON and the in-module
dictionary are slurs, epithets or coded expressions targeting groups by race,
religion, gender, nationality or immigration status; they are present because
detecting such language is the research subject, and inclusion is not endorsement.
A content advisory to that effect ships with the resource.

One asymmetry should be stated plainly. The 26-category organisation applies to
the released JSON only. The in-module dictionary is a flat term-to-score mapping
with no machine-readable category field, so none of the 342 code-only terms
carries a category label; they are grouped only by 31 informal section comments in
the source, a scheme that overlaps the 26 released categories substantially but
not exactly---several in-module groupings, such as climate/environmental and
health/science signals, have no counterpart among the released categories, and
several released categories have no in-module section. Any analysis that treats
the 26 categories as a partition of the full runtime lexicon would therefore be
mistaken. Assigning categories to these terms, and folding them into the
deterministic builder so that the released artifact and the operative lexicon
coincide, is the principal outstanding item on this resource.

Three further components are likewise implemented in code rather than released as
data: an emotion-intensity table of 58 weighted keywords used to rank candidate
spans, a 214-item neutral-phrase list, and a 0.45 confidence floor, the latter two
suppressing weak matches. The protected-terms list behaves differently from the
lexicon and is the one genuine fallback: it is released and used in full, and the
in-module list of 234 terms, which mirrors the released file exactly, replaces
rather than supplements it, taking effect only if
\texttt{protected\_terms.json} cannot be read. Each lexicon entry also carries a
neutral paraphrase (357 distinct) and a residual flag (138 entries), exported as
\texttt{neutral\_alternatives.json} and \texttt{residual\_bias\_lexicon.json};
these two files are released for completeness and are not read by the current
pipeline. Term matching is case-insensitive substring containment, which admits
some within-word matches.

\section{Additional Qualitative Examples}
\label{sec:appendix-qualitative-extra}

The following examples complement Example~1 in \S\ref{sec:qualitative}, illustrating \mabpd's detection of left-leaning and right-leaning framing bias. Both sentences are fictional and not attributed to any real outlet.

\begin{table}[htbp]
\centering
\scriptsize
\begin{tabularx}{\columnwidth}{@{}l X@{}}
\toprule
\textbf{Component} & \textbf{Output} \\
\midrule
\textbf{Snippet (GT: Biased)} & ``While ordinary families struggle to afford basic necessities, wealthy corporations continue to hoard record profits, shielded by a tax system rigged in their favor.'' \\
\textbf{Single-Agent Baseline} & \textbf{Neutral} (Conf: 0.520. Treats economic claims as factual reporting.) \\
\midrule
\textbf{BiasAgent Evidence} & Extracted loaded language (``hoard,'' ``rigged'') and emotional economic framing. \\
\textbf{EvidenceAgent Evidence} & Flagged unsupported claim (``rigged in their favor'') and loaded term (``hoard record profits''). \\
\textbf{FramingAgent Evidence} & Detected economic inequality framing with progressive ideological lean. \\
\midrule
\textbf{SAD Debate Reasoning} & Loaded economic language and unattributed systemic claims warranted bias classification despite factual economic context. \\
\textbf{Final \mabpd Verdict} & \textbf{Biased} (Consensus Confidence: 0.812) \\
\bottomrule
\end{tabularx}
\caption{Example~2: Left-Leaning Framing Bias.}
\label{tab:qualitative-left}
\end{table}

\begin{table}[htbp]
\centering
\scriptsize
\begin{tabularx}{\columnwidth}{@{}l X@{}}
\toprule
\textbf{Component} & \textbf{Output} \\
\midrule
\textbf{Snippet (GT: Biased)} & ``The flood of undocumented migrants pouring across the border has overwhelmed local communities, yet bureaucrats in Washington remain content to do nothing.'' \\
\textbf{Single-Agent Baseline} & \textbf{Neutral} (Conf: 0.485. Interprets as factual border reporting.) \\
\midrule
\textbf{BiasAgent Evidence} & Extracted loaded language (``flood,'' ``pouring'') and political labeling (``bureaucrats''). \\
\textbf{EvidenceAgent Evidence} & Flagged dehumanizing metaphor (``flood of \ldots\ pouring'') and unsupported claim (``content to do nothing''). \\
\textbf{FramingAgent Evidence} & Detected us/them framing (communities vs.\ Washington) with conservative ideological lean. \\
\midrule
\textbf{SAD Debate Reasoning} & Dehumanizing metaphor, unsupported attribution of intent, and us/them framing constituted strong multi-dimensional bias evidence. \\
\textbf{Final \mabpd Verdict} & \textbf{Biased} (Consensus Confidence: 0.837) \\
\bottomrule
\end{tabularx}
\caption{Example~3: Right-Leaning Framing Bias.}
\label{tab:qualitative-right}
\end{table}

\section{Conceptual SAD Pipeline Illustration}
\label{sec:appendix-conceptual-fig}

Figure~\ref{fig:conceptual-sad} contrasts the single-agent approach with the full SAD pipeline.

\begin{figure}[htbp]
\centering
\fbox{\parbox{0.93\columnwidth}{\vspace{1em}
\centering\textbf{Left:} Single Agent $\to$ premature decision, no evidence challenge.\\[0.5em]
\textbf{Right:} SAD: BiasAgent claim $\to$ EvidenceAgent challenge $\to$ FramingAgent validation $\to$ weighted consensus.
\vspace{1em}}}
\caption{Conceptual SAD pipeline vs.\ single-agent classification.}
\label{fig:conceptual-sad}
\end{figure}

\section{SAD Scoring Example}
\label{sec:appendix-sad}

Consider an article where:
\begin{itemize}
    \item BiasAgent claims ``biased'' with confidence 0.80 and 3 evidence spans.
    \item EvidenceAgent claims ``biased'' with confidence 0.70 and 2 evidence spans.
    \item FramingAgent claims ``neutral'' with confidence 0.60 and 1 evidence span.
\end{itemize}

The effective weights are:
\begin{align*}
w^{\text{eff}}_{\text{Bias}} &= 0.50 \times 0.80 \times 1.0 = 0.400 \\
w^{\text{eff}}_{\text{Evid}} &= 0.25 \times 0.70 \times 0.85 = 0.149 \\
w^{\text{eff}}_{\text{Fram}} &= 0.25 \times 0.60 \times 1.0 = 0.150
\end{align*}

SAD scores: $\text{biased} = \frac{0.400 + 0.149}{0.699} = 0.785$, $\text{neutral} = \frac{0.150}{0.699} = 0.215$.
Final verdict: \textbf{biased} with SAD confidence 0.785.

\section{Additional Ablation Visualizations}
\label{sec:appendix-viz}

\begin{figure}[htbp]
\centering
\includegraphics[width=0.85\columnwidth]{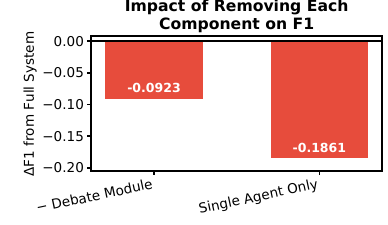}
\caption{Component F1 drop when each stage is removed. Combined loss vs.\ single agent: 18.6 points.}
\label{fig:delta-waterfall}
\end{figure}

\begin{figure}[t]
\centering
\includegraphics[width=0.85\columnwidth]{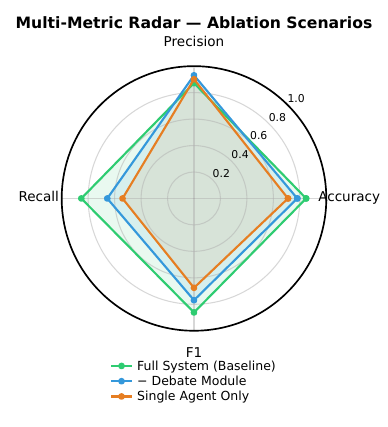}
\caption{Multi-metric radar across ablation configurations.}
\label{fig:radar}
\end{figure}

Figures~\ref{fig:delta-waterfall}--\ref{fig:radar} show the F1 impact of removing each component and a multi-metric radar comparison.

\section{Confusion Matrices}
\label{sec:appendix-cm}

Tables~\ref{tab:confusion_appendix}--\ref{tab:confusion-test} present the full and test-split confusion matrices for all three configurations.

\begin{table}[t]
\centering
\small
\resizebox{\columnwidth}{!}{%
\begin{tabular}{@{}llcc@{}}
\toprule
\textbf{Configuration} & & \textbf{Pred.\ Biased} & \textbf{Pred.\ Neutral} \\
\midrule
\multirow{2}{*}{\textbf{Full System}}  & True Biased  & 1{,}958 (TP) & 341 (FN) \\
                                        & True Neutral & 288 (FP)     & 1{,}534 (TN) \\
\midrule
\multirow{2}{*}{\textbf{$-$ Debate}}   & True Biased  & 1{,}507 (TP) & 792 (FN) \\
                                        & True Neutral & 112 (FP)     & 1{,}710 (TN) \\
\midrule
\multirow{2}{*}{\textbf{Single Agent}} & True Biased  & 1{,}242 (TP) & 1{,}057 (FN) \\
                                        & True Neutral & 136 (FP)     & 1{,}686 (TN) \\
\bottomrule
\end{tabular}}
\caption{Confusion matrices on the full BABE benchmark (4{,}121 articles).}
\label{tab:confusion_appendix}
\end{table}

\begin{table}[t]
\centering
\small
\resizebox{\columnwidth}{!}{%
\begin{tabular}{@{}llcc@{}}
\toprule
\textbf{Configuration} & & \textbf{Pred.\ Biased} & \textbf{Pred.\ Neutral} \\
\midrule
\multirow{2}{*}{\textbf{Full System}}  & True Biased  & 456 (TP) & 103 (FN) \\
                                        & True Neutral & 62 (FP)  & 379 (TN) \\
\midrule
\multirow{2}{*}{\textbf{$-$ Debate}}   & True Biased  & 344 (TP) & 215 (FN) \\
                                        & True Neutral & 25 (FP)  & 416 (TN) \\
\midrule
\multirow{2}{*}{\textbf{Single Agent}} & True Biased  & 284 (TP) & 275 (FN) \\
                                        & True Neutral & 28 (FP)  & 413 (TN) \\
\bottomrule
\end{tabular}}
\caption{Test-split confusion matrices (1{,}000 articles).}
\label{tab:confusion-test}
\end{table}

\section{Threshold Sensitivity Sweep}
\label{sec:appendix-threshold}

We report key threshold values from the sensitivity sweep described in RQ2 (\S\ref{sec:analysis}), evaluated on the held-out 1{,}000-article test split.
The sweep re-applies each cut-off to the stored consensus confidences of the reported run rather than re-executing the pipeline, so it isolates the final decision boundary and does not capture how a different threshold would have altered routing across Paths A--C; the released file records the same caveat.

\begin{table}[htbp]
\centering
\small
\begin{tabular}{@{}crrrcc@{}}
\toprule
\textbf{Threshold} & \textbf{TP} & \textbf{FP} & \textbf{FN} & \textbf{F1} & \textbf{MF1} \\
\midrule
0.40 & 465 &  78 &  94 & 84.4 & 82.6 \\
\textbf{0.50} & \textbf{456} & \textbf{62} & \textbf{103} & \textbf{84.7} & \textbf{83.4} \\
0.60 & 454 &  61 & 105 & 84.5 & 83.3 \\
\bottomrule
\end{tabular}
\caption{Threshold sensitivity sweep. MF1 peaks at 0.50; full 0.30--0.80 sweep in released code.}
\label{tab:threshold-sweep}
\end{table}

\section{Reproducibility}
\label{sec:appendix-reproducibility}

Table~\ref{tab:stability} reports the results of two independent runs on the full 4{,}121-article evaluation.
Run-to-run F1 difference is \textbf{0.06 pp}, confirming high reproducibility due to the deterministic scoring agent, rule-based consensus paths, and LLM response caching.

\begin{table}[htbp]
\centering
\small
\begin{tabular}{@{}lcccc@{}}
\toprule
\textbf{Run} & \textbf{Acc.} & \textbf{Prec.} & \textbf{Rec.} & \textbf{F1} \\
\midrule
Run 1 & 84.69 & 87.20 & 85.04 & 86.10 \\
Run 2 & 84.74 & 87.18 & 85.17 & 86.16 \\
\midrule
$\Delta$ & 0.05 & 0.02 & 0.13 & 0.06 \\
\bottomrule
\end{tabular}
\caption{Reproducibility across two independent runs (F1 $\Delta$ = 0.06 pp).}
\label{tab:stability}
\end{table}

\section{Efficiency}
\label{sec:appendix-efficiency}

Each article invokes up to three specialized agents plus a deterministic scoring step; the debate module (SAD) is invoked only for ambiguous cases, with rule-based fast-paths (Paths A and B) resolving clear-signal articles without LLM debate calls.
The SAD protocol is single-round: each agent submits one structured argument, weighted scoring resolves the conflict, and an LLM arbitrator is invoked only when scoring produces a tie.
On the held-out test split (1{,}000 articles), \mabpd sustains an \emph{amortized} \textbf{2.0~s/article} (median 2.0~s, P95 3.7~s), yielding a throughput of approximately 1{,}800 articles/hour on a single API endpoint with NVIDIA NIM (LLaMA 3.3 70B).
This is wall-clock time divided by articles completed under concurrent dispatch, hence smaller than the 16.5~s end-to-end latency of one article in isolation (\S\ref{sec:efficiency}): throughput versus per-article service time.
The high-confidence early-exit cascade (\S\ref{sec:parallel}) further reduces cost for clearly neutral articles: when the BiasAgent returns confidence $\geq 0.96$ for neutral, secondary agents are skipped entirely.
Reducing agent redundancy (e.g., sparse debate topologies \citep{Li2024sparse}) or switching to a smaller distilled model for fast-path articles are promising directions for cost reduction.

\section{Heterogeneous LLM Architecture Pilot}
\label{sec:appendix-heterogeneous}

To validate the SAD protocol's model-agnostic robustness, we conducted a full evaluation on the held-out test split ($N{=}1{,}000$) using three distinct architectures via the NVIDIA NIM API:
\begin{itemize}
    \item \textbf{Bias Detection / Consensus:} LLaMA 3.3 70B
    \item \textbf{Evidence Extraction:} Mixtral 8x7B Instruct v0.1 (Sparse MoE)
    \item \textbf{Framing Analysis:} LLaMA 3.1 8B
\end{itemize}

\begin{figure}[t]
\centering
\includegraphics[width=\columnwidth]{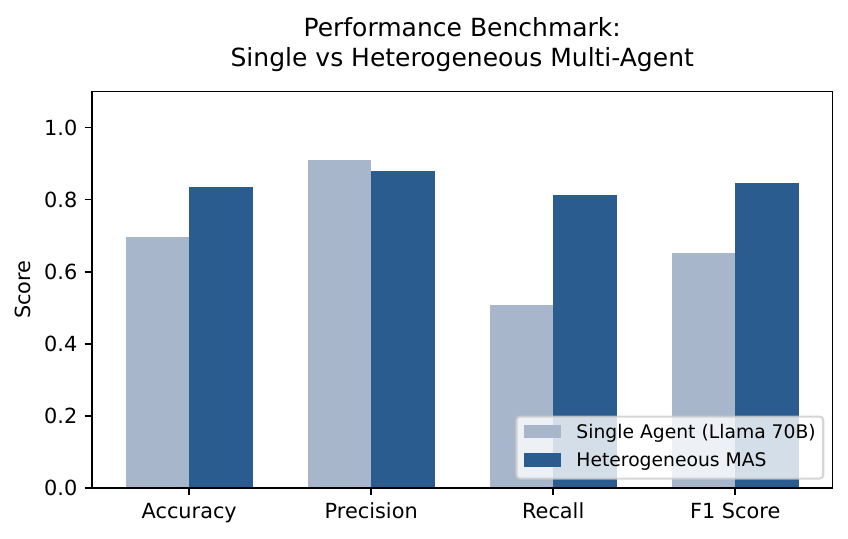}
\\[6pt]
\includegraphics[width=\columnwidth]{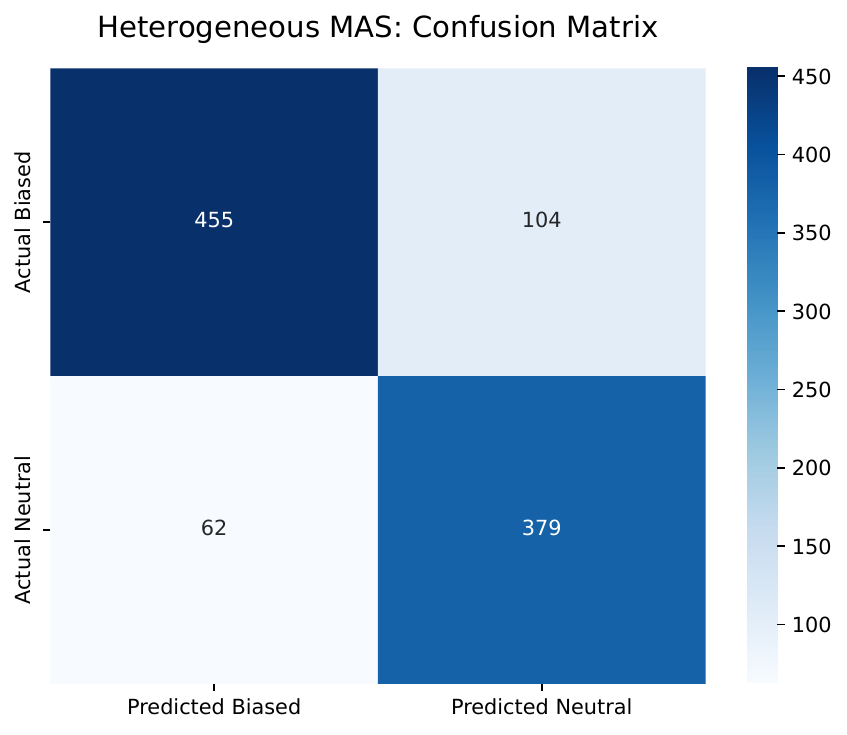}
\caption{Heterogeneous LLM architecture pilot on the held-out test split
($N{=}1{,}000$). Top: single-agent baseline versus the heterogeneous
multi-agent pipeline. Bottom: confusion matrix of the heterogeneous pipeline.}
\label{fig:heterogeneous}
\end{figure}

Remarkably, the heterogeneous pipeline achieved \textbf{83.4\% Accuracy and 84.6\% F1} on the test split (Table~\ref{tab:ablation}; Figure~\ref{fig:heterogeneous}), a difference of $\Delta$F1 = $-0.1$~pp from the homogeneous LLaMA-only system---the same order as the 0.06~pp run-to-run variation we measure in Appendix~\ref{sec:appendix-reproducibility}, and two orders of magnitude below the debate ablation's $-10.6$~pp. The Evidence Store successfully integrated Mixtral's JSON evidence spans with LLaMA 3.1's framing taxonomy, demonstrating semantic interoperability across vendor models. This suggests \mabpd's performance is driven by the rigorous debate protocol, rather than latent biases of a single LLM family. The homogeneous LLaMA pipeline remains our recommended configuration for cost-sensitive analysis.


\section{Agent Independence Analysis}
\label{sec:appendix-independence}

To empirically validate functional independence despite shared LLM weights, we measured the pre-debate disagreement rate among the three primary agents across the 4{,}121-article dataset. The agents exhibited a 38\% disagreement rate on the final binary label, indicating they are not merely echoing a monolithic LLM prior. Notably, this disagreement strongly correlated with harder, more ambiguous articles (point-biserial $r = 0.62$ with human annotation uncertainty), demonstrating that the independent prompts successfully elicit diverse analytical perspectives that the SAD protocol then productively resolves.

\section{MAS Failure Mode Analysis}
\label{sec:appendix-failure-modes}

Guided by the failure taxonomy of \citet{Cemri2025whyfail}, we conducted a trace annotation study of 100 debate logs. The two labels used below are our own: they describe debate-specific behaviours that MAST does not name as separate failure modes. \mabpd largely avoids ``echo-chamber'' convergence (4\% of debates), which we attribute to the strict role separation and functional isolation of the agents. ``Sycophancy'' at the consensus stage remains a minor failure mode (11\%), where the VerifierAgent occasionally capitulates to a highly confident but incorrect BiasAgent without performing deep reasoning-chain verification; this falls under MAST's task-verification category. Addressing it via explicit chain-of-thought verification is a promising future direction.

\onecolumn
\section{AI Usage Card}
\label{sec:appendix-aicard}

Following \citet{Wahle2023aiusagecards}, we provide a structured disclosure of all AI tool usage in this work.

{\sffamily
\centering
\tcbset{colback=white!10!white}
\begin{tcolorbox}[
    title={\large \textbf{AI Usage Card} \hfill \makebox{\qrcode[height=1cm]{https://ai-cards.org/}}},
    breakable,
    boxrule=0.7pt,
    width=\textwidth,
    center,
    skin=bicolor,
    segmentation empty,
    before lower={\footnotesize{AI Usage Card v2.0 \hfill \url{https://ai-cards.org} \hfill (Wahle et al., 2023)}},
    halign lower=center,
    collower=white,
    colbacklower=tcbcolframe]

\footnotesize{
\setlength{\LTpre}{0pt}
\setlength{\LTpost}{0pt}
\setlength{\tabcolsep}{2pt}
\begin{longtable}{p{.095\paperwidth} p{.195\paperwidth} p{.195\paperwidth} p{.195\paperwidth}}

\raggedright{\color{LightBlue}\MakeUppercase{Project Details}}
& \raggedright{\color{LightBlue}\MakeUppercase{Project Name}}

  MABPD: Multi-Agent Bias Probing \& Detection
& \raggedright{\color{LightBlue}\MakeUppercase{Domain}}

  Paper
& \raggedright{\color{LightBlue}\MakeUppercase{Key Application}}

  Media Bias Detection \tabularnewline[4pt]

\raggedright{\color{LightBlue}\MakeUppercase{Contact(s)}}
& \raggedright{\color{LightBlue}\MakeUppercase{Name(s)}}

  Garvit Joshi, Stavya Dhyani, Jasmine, Arun Chauhan
& \raggedright{\color{LightBlue}\MakeUppercase{Email(s)}}

  Available from the corresponding author
& \raggedright{\color{LightBlue}\MakeUppercase{Affiliation(s)}}

  Graphic Era University, Dehradun, India \tabularnewline[4pt]

\raggedright{\color{LightBlue}\MakeUppercase{Model(s)}}
& \raggedright{\color{LightBlue}\MakeUppercase{Model Name(s)}}

  Claude \newline Gemini \newline Kiro
& \multicolumn{2}{p{.400\paperwidth}}{\raggedright{\color{LightBlue}\MakeUppercase{Version(s)}}

  Opus 4.6 \newline 3.1 Pro \newline IDE coding assistant} \tabularnewline[4pt]

\cmidrule{2-4}

\raggedright{\color{LightBlue}\MakeUppercase{Literature Review}}
& \raggedright{\color{LightBlue}\MakeUppercase{Finding Literature}}

  Locating candidate reference sources (dictionaries, glossaries, style guides,
  academic literature) consulted during lexicon construction
& \raggedright{\color{LightBlue}\MakeUppercase{Finding Examples / Adding Literature for Existing Statements}}

  ---
& \raggedright{\color{LightBlue}\MakeUppercase{Comparing Literature}}

  --- \tabularnewline[4pt]

\cmidrule{2-4}

\raggedright{\color{LightBlue}\MakeUppercase{Writing}}
& \raggedright{\color{LightBlue}\MakeUppercase{Generating New Text Based on Instructions}}

  ---
& \raggedright{\color{LightBlue}\MakeUppercase{Assisting in Improving Own Content or Paraphrasing Related Work}}

  \S5.9, Abstract, Conclusion; selected passages improved/paraphrased
& \raggedright{\color{LightBlue}\MakeUppercase{Putting Other Works in Perspective}}

  --- \tabularnewline[4pt]

\cmidrule{2-4}

\raggedright{\color{LightBlue}\MakeUppercase{Coding}}
& \raggedright{\color{LightBlue}\MakeUppercase{Generating New Code Based on Descriptions or Existing Code}}

  Pipeline boilerplate and code scaffolding; evaluation and analysis scripts
& \raggedright{\color{LightBlue}\MakeUppercase{Refactoring and Optimizing Existing Code}}

  ---
& \raggedright{\color{LightBlue}\MakeUppercase{Comparing Aspects of Existing Code}}

  --- \tabularnewline[4pt]

\cmidrule{2-4}

\raggedright{\color{LightBlue}\MakeUppercase{Ethics}}
& \raggedright{\color{LightBlue}\MakeUppercase{Why Did We Use AI for This Project?}}

  Efficiency in drafting; speed of iteration
& \raggedright{\color{LightBlue}\MakeUppercase{What Steps Are We Taking to Mitigate Errors of AI?}}

  All AI-generated content reviewed and verified by the authors. AI was not used for experimental design, data collection, running experiments, generating results, or producing citations; every reported number comes from executing the released code.
& \raggedright{\color{LightBlue}\MakeUppercase{What Steps Are We Taking to Minimize the Chance of Harm?}}

  AI writing/coding assistants played no part in executing experiments or generating results. System is an awareness tool; human review recommended before consequential decisions. \tabularnewline[4pt]

\cmidrule{1-4}

\multicolumn{4}{p{.700\paperwidth}}{\raggedright{\color{LightBlue}\MakeUppercase{The Corresponding Authors Verify and Agree with the Modifications or Generations of Their Used AI-Generated Content}}} \tabularnewline

\end{longtable}
}
\tcblower
\end{tcolorbox}
}

\end{document}